\documentclass{l4dc2026}

\usepackage{amsmath,amsfonts,bm,mathtools}
\usepackage{mathrsfs}
\usepackage{pgf,tikz}
\usetikzlibrary{positioning}
\usepackage{pgfplots}
\pgfplotsset{compat = newest}
\pgfplotsset{every x tick label/.append style={font=\footnotesize, yshift=0.5ex}}
\pgfplotsset{every y tick label/.append style={font=\footnotesize, xshift=0.5ex}}

\def\eqref#1{equation~\ref{#1}}

\def\1{\bm{1}}

\DeclareMathAlphabet{\mathsfit}{\encodingdefault}{\sfdefault}{m}{sl}
\SetMathAlphabet{\mathsfit}{bold}{\encodingdefault}{\sfdefault}{bx}{n}

\def\gW{{\mathcal{W}}}

\def\sW{{\mathbb{W}}}

\DeclareMathOperator{\sign}{sign}

\newcommand{\trace}[1]{\mbox{trace}\left(#1\right)}

\newcommand{\Reals}[1][\empty]{\mathbb{R}^{#1}}

\newcommand{\SSp}{\sW}

\newcommand{\TSet}{\mathscr{T}}
\newcommand{\CSetf}{\mathscr{G}_f}

\newcommand{\cost}{f}
\newcommand{\costovp}{g}
\newcommand{\fmin}{\underline{\cost}}

\newcommand{\bfW}{\mathbf{W}}
\newcommand{\bfw}{\mathbf{w}}

\newcommand{\dd}{\mathrm{d}}

\newcommand{\PLI}{P\L{}I\;}
\newcommand{\pdPLI}{\textit{pd}\PLI}

\tikzstyle{neuron} = [draw, fill=white, circle, 
    minimum height=3.5em, minimum width=1em, style=thick]
\allowdisplaybreaks

\title[On the Convergence of Neural Networks]{On the Convergence of Overparameterized Problems: \\Inherent Properties of the Compositional Structure of Neural Networks}
\usepackage{times}

\coltauthor{\Name{Arthur Castello B. de {Oliveira}} \Email{a.castello@northeastern.edu}\\
 \Name{Dhruv D. Jatkar} \Email{jatkar.d@northeastern.edu}\\
 \Name{Eduardo D. {Sontag}} \Email{e.sontag@northeastern.edu}\\
 \addr Northeastern University, 805 Columbus Ave, Boston, MA 02120}

\begin{document}

\maketitle

\begin{abstract}%
 This paper investigates how the compositional structure of neural networks shapes their optimization landscape and training dynamics. We analyze the gradient flow associated with overparameterized optimization problems, which can be interpreted as training a neural network with linear activations. Remarkably, we show that the global convergence properties can be derived for any cost function that is proper and real analytic. We then specialize the analysis to scalar-valued cost functions, where the geometry of the landscape can be fully characterized. In this setting, we demonstrate that key structural features -- such as the location and stability of saddle points -- are universal across all admissible costs, depending solely on the overparameterized representation rather than on problem-specific details. Moreover, we show that convergence can be arbitrarily accelerated depending on the initialization, as measured by an imbalance metric introduced in this work. Finally, we discuss how these insights may generalize to neural networks with sigmoidal activations, showing through a simple example which geometric and dynamical properties persist beyond the linear case.
\end{abstract}


\section{Introduction}

    The widespread success of artificial intelligence (AI) in solving complex tasks has ignited interest in understanding the theoretical underpinnings of modern learning systems \cite{belkin2019reconciling,bartlett2020benign,arora2019implicit,du2019gradient}. 
    
    Among the many approaches to this question, a particularly fruitful direction studies linear neural networks as tractable models for analyzing how the compositional structure of deep networks affects their optimization landscape and training dynamics. The simplicity of linear activations allows one to isolate the geometric and dynamical consequences of composition itself, independent of nonlinearities \cite{bah_learning_2022,arora2019implicit,kawaguchi_deep_2016,chitour2018geometric,min2021explicit,menon2023geometry,min2023convergence,de2023dynamics,de2024remarks,de2025convergence}.

    A rich body of work has characterized the behavior of linear networks in regression and related problems, revealing elegant structures in their loss landscapes and convergence dynamics \cite{bah_learning_2022,kawaguchi_deep_2016,arora2019implicit,chitour2018geometric,min2021explicit}. More recent research has extended these results to broader classes of cost functions \cite{menon2023geometry,min2023convergence} and to nonconvex, control-related optimization problems \cite{de2025convergence}. Notably, \cite{min2023convergence} demonstrates an initialization-dependent exponential acceleration of gradient-flow solutions for strongly convex objectives, indicating a possible advantage of adopting an overparameterized formulation over regular gradient flow. In \cite{de2025convergence} we further explored this property in a problem that is non-convex and not gradient dominant -- the policy optimization problem for the linear quadratic regulator (LQR). Furthermore, in the same paper we showed that all the good properties previously demonstrated for the specific case of linear regression also hold for this more complex nonconvex problem. Building on these observations, in \cite{wafi2025almost} we later demonstrated that overparameterization induces not merely quantitative acceleration but also qualitative improvements in the convergence profile of gradient-flow solutions.

    The present paper seeks to determine to what extent these convergence properties are inherent to the compositional structure of overparameterization itself, rather than to specific features of the underlying cost function. We show that for any proper, real-analytic cost, the corresponding overparameterized gradient flow -- interpreted as training a deep linear network -- admits invariant quantities that partition the parameter space into disjoint invariant manifolds. Through this geometric property we prove that gradient-flow trajectories always converge to a critical point of the problem (despite the overparameterized cost not being proper anymore), and almost everywhere to critical points of the original problem if the linear neural network has a single hidden layer. For scalar costs, we prove that the geometry of the center–stable manifolds introduced by overparameterization is universal, depending only on the compositional structure and not on the specific cost, while the rate of convergence depends on an initialization-imbalance measure introduced here.
    
    This paper is organized as follows. We begin in Section \ref{sec:background} by formally defining the class of optimization problems under scrutiny, and the exact structure of an overparameterized formulation. We then discuss the invariant measure, how it is a consequence of the compositional structure, and why it is important for studying the convergence of gradient-flow solutions. This result allows us to state the two main results regarding convergence of solutions for overparameterized problems. Next we move to scalar optimization problems in Section \ref{sec:vectorcase} proving the aforementioned problem-independent structure of the parameter space and the initialization-based acceleration of solutions. We then explore in Section \ref{sec:sigmoidalextension} how the results of this paper might be extended to nonlinear neural networks, proving, for one example, the existence of an invariant quantity even for this nonlinear case. Finally, in Section \ref{sec:conclusion} we review the results of the paper and their importance for understanding the training behavior of neural networks. Most proofs are deferred to the Appendix, but each result is followed by a sketch of its proof.

\section{Overparameterization and neural networks}
\label{sec:background}

    Let $\cost:\SSp\to\Reals$ be a proper (i.e. the preimage of each compact set is compact), bounded below, continuously differentiable function with a Lipschitz gradient, where $\SSp\subseteq\Reals[n\times n]$ is a simply connected set.  For such $\cost$, define an optimization problem as
    \begin{align}
        \label{eq:opt-problem}
        \underset{W\in\SSp}{\text{minimize}} \quad &\cost(W).
    \end{align}
    Assume further that the minimum value of the function exists and is given by $\fmin:=\inf_{W\in\SSp}f(W)$, and solving \eqref{eq:opt-problem} means finding any point $W^*\in\TSet:=\{{W}\in\SSp~|~f({W})=\fmin\}$. 
    
    Gradient methods are a popular approach to finding a candidate solution to \eqref{eq:opt-problem}. Specifically, a gradient-flow solution consists of solving the following initial value problem
    \begin{equation}
        \label{eq:gradflow-def}
        \dot W = -\nabla f(W), \quad W(0)=W_0
    \end{equation}
    for some $W_0\in\SSp$. It is easy to verify that properness of $f$ implies pre-compactness of solutions of \eqref{eq:gradflow-def}, which by the Krasovskii-LaSalle's principle implies convergence of solutions to $\CSetf:=\{W\in\SSp~|~\nabla f(W)=0\}$. Furthermore, one can verify that $\TSet\subseteq\CSetf$, but $\CSetf\not\subseteq\TSet$ in general, and thus further assumptions are often required in practice. To simplify the analysis in this paper, we also assume that for all $W\in\CSetf$, $\mbox{rank}{(W)}=n$.

    For a given width $k$ and depth $N$, let $k>n$, $W_1\in\Reals[k\times n]$, $W_N\in\Reals[n\times k]$ and $W_i\in\Reals[k\times k]$. Then we define an overparameterized formulation of \eqref{eq:opt-problem} as
    %
        \begin{align}
            \label{eq:ovp-problem}
        \underset{(W_1,\dots,W_N)\in\SSp}{\text{minimize}} \quad &\costovp(W_1,\dots,W_N),
        \end{align}
    %
    where $\costovp(W_1,\dots,W_N):=\cost(W_{N} W_{N-1}\cdots W_2 W_1)$ and $(W_1,\dots,W_N)\in\SSp$ is an abuse of notation to mean $$(W_1,\dots,W_N)\in\left\{(W_1,\dots,W_N)\in\Reals[k\times n]\times\left(\Reals[k\times k]\right)^{N-2}\times\Reals[n\times k]~\big|~W_N W_{N-1}\cdots W_2 W_1\in\SSp\right\}.$$
    Notice that despite $f$ being a proper function, $g$ is not. This happens because for any $\eta\in\mathbb{R}$, $g(W_1,W_2)=g(\eta W_1,(1/\eta)W_2)$ which introduces unbounded directions in the state space for which the cost does not explode depite the norm of the parameters exploding.

    \sloppy For simplicity of notation define $\bfW(W):=W_N W_{N-1}\cdots W_2 W_1$ for any given $W=(W_1,\dots,W_N)$, where the dependency of $\bfW$ on $W$ can be omitted if it is clear from context.

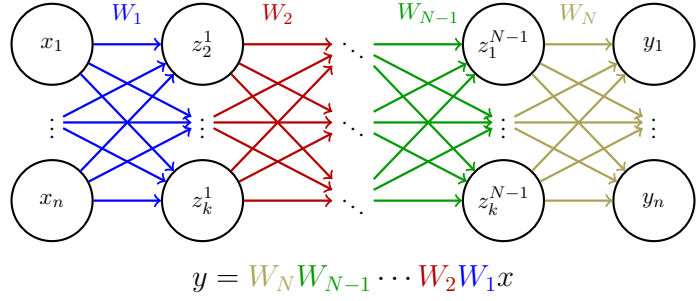
\begin{figure}
        \centering
        \begin{tikzpicture}[scale=0.8,transform shape,node distance=1.3cm]
            \node [neuron, name=neui1] { $x_1$};
            \node [rectangle, name=neui2, below of = neui1] { \vdots};
            \node [neuron, name=neui3, below of = neui2] { $x_n$};
            
            \node [rectangle, name=neuh13, right of = neui2, node distance = 2.5cm] { \vdots};
            \node [neuron, name=neuh12, above of = neuh13] { $z^1_2$};
            \node [neuron, name=neuh14, below of = neuh13] { $z^1_{k}$};
    
            \node [rectangle, name=neud1, right of = neuh13, node distance = 2.5cm] {$\mathbf{\ddots}$};
            \node [rectangle, name=neud2, above of = neud1] {$\mathbf{\ddots}$};
            \node [rectangle, name=neud3, below of = neud1] {$\mathbf{\ddots}$};

            \node[rectangle,name=equation,below of =neud3]{\Large $y=\textcolor{yellow!60!black}{W_N}\textcolor{green!60!black}{W_{N-1}}\cdots \textcolor{red!70!black}{W_2}\textcolor{blue}{W_1}x$};
    
            \node [rectangle, name=neuhN3, right of = neud1, node distance = 2.5cm] { \vdots};
            \node [neuron, name=neuhN2, above of = neuhN3] { $z^{N-1}_1$};
            \node [neuron, name=neuhN1, below of = neuhN3] { $z^{N-1}_{k}$};
    
            \node [rectangle, name=neuy1, right of = neuhN3, node distance = 2.5cm] {\vdots};
            
            \node [neuron, name=neuy2, above of = neuy1] { $y_1$};
            \node [neuron, name=neuy3, below of = neuy1] { $y_n$};
    
            \node [rectangle, name=aux1, right of = neui1, node distance = 1.25cm] {};
            \node [rectangle, name=K1, above of = aux1, node distance = 0.5cm] {\color{blue}$W_1$};
            \node [rectangle, name=K2, right of = K1, node distance = 2.5cm] {\color{red!70!black}$W_2$};
            \node [rectangle, name=KN1, right of = K2, node distance = 2.5cm] {\color{green!60!black}$W_{N-1}$};
            \node [rectangle, name=KN, right of = KN1, node distance = 2.5cm] {\color{yellow!60!black}$W_N$};
    
            \draw [->, line width=0.3mm, color=blue] (neui1) -- (neuh12) {};
            \draw [->, line width=0.3mm, color=blue] (neui1) -- (neuh13) {};
            \draw [->, line width=0.3mm, color=blue] (neui1) -- (neuh14) {};
            
            \draw [->, line width=0.3mm, color=blue] (neui2) -- (neuh12) {};
            \draw [->, line width=0.3mm, color=blue] (neui2) -- (neuh13) {};
            \draw [->, line width=0.3mm, color=blue] (neui2) -- (neuh14) {};
            
            \draw [->, line width=0.3mm, color=blue] (neui3) -- (neuh12) {};
            \draw [->, line width=0.3mm, color=blue] (neui3) -- (neuh13) {};
            \draw [->, line width=0.3mm, color=blue] (neui3) -- (neuh14) {};
            
            \draw [->, line width=0.3mm, color=red!70!black] (neuh12) -- (neud1) {};
            \draw [->, line width=0.3mm, color=red!70!black] (neuh13) -- (neud1) {};
            \draw [->, line width=0.3mm, color=red!70!black] (neuh14) -- (neud1) {};
    
            \draw [->, line width=0.3mm, color=red!70!black] (neuh12) -- (neud2) {};
            \draw [->, line width=0.3mm, color=red!70!black] (neuh13) -- (neud2) {};
            \draw [->, line width=0.3mm, color=red!70!black] (neuh14) -- (neud2) {};
    
            \draw [->, line width=0.3mm, color=red!70!black] (neuh12) -- (neud3) {};
            \draw [->, line width=0.3mm, color=red!70!black] (neuh13) -- (neud3) {};
            \draw [->, line width=0.3mm, color=red!70!black] (neuh14) -- (neud3) {};
    
            \draw [->, line width=0.3mm, color=green!60!black] (neud1) -- (neuhN1) {};
            \draw [->, line width=0.3mm, color=green!60!black] (neud1) -- (neuhN2) {};
            \draw [->, line width=0.3mm, color=green!60!black] (neud1) -- (neuhN3) {};
    
            \draw [->, line width=0.3mm, color=green!60!black] (neud2) -- (neuhN1) {};
            \draw [->, line width=0.3mm, color=green!60!black] (neud2) -- (neuhN2) {};
            \draw [->, line width=0.3mm, color=green!60!black] (neud2) -- (neuhN3) {};
    
            \draw [->, line width=0.3mm, color=green!60!black] (neud3) -- (neuhN1) {};
            \draw [->, line width=0.3mm, color=green!60!black] (neud3) -- (neuhN2) {};
            \draw [->, line width=0.3mm, color=green!60!black] (neud3) -- (neuhN3) {};
    
             \draw [->, line width=0.3mm, color=yellow!60!black] (neuhN1) -- (neuy1) {};
            \draw [->, line width=0.3mm, color=yellow!60!black] (neuhN2) -- (neuy1) {};
            \draw [->, line width=0.3mm, color=yellow!60!black] (neuhN3) -- (neuy1) {};
    
            \draw [->, line width=0.3mm, color=yellow!60!black] (neuhN1) -- (neuy2) {};
            \draw [->, line width=0.3mm, color=yellow!60!black] (neuhN2) -- (neuy2) {};
            \draw [->, line width=0.3mm, color=yellow!60!black] (neuhN3) -- (neuy2) {};
    
            \draw [->, line width=0.3mm, color=yellow!60!black] (neuhN1) -- (neuy3) {};
            \draw [->, line width=0.3mm, color=yellow!60!black] (neuhN2) -- (neuy3) {};
            \draw [->, line width=0.3mm, color=yellow!60!black] (neuhN3) -- (neuy3) {};
        \end{tikzpicture}
        \caption{Depiction of a linear neural network.}
        \label{fig:LFFNN}
    \end{figure}

    Notice that $\bfW$ is the resulting expression of a feedforward neural network with weight matrices $W$, linear activations, $N-1$ hidden layers, and width of $k$. Fig. \ref{fig:LFFNN} illustrates this interpretation, showing how the compositional structure of feedforward neural networks result in the proposed overparameterized formulation.

    This is a well-studied formulation in the literature \cite{bah_learning_2022,chitour2018geometric,de2023dynamics,de2024remarks,eftekhari_training_2020,kawaguchi_deep_2016,min2021explicit,min2023convergence}, with many results characterizing the optimization landscape for different applications, as well as providing guarantees for gradient methods. An \emph{overparameterized gradient flow} is defined as a set of ODEs given by
    \begin{equation}
        \label{eq:gradflowOVP-def}
        \dot W_i:=-\nabla_{W_i}g(W_1,\dots,W_N)=-(W_N\dots W_{i+1})^\top\nabla f(\bfW)(W_{i-1}\dots W_1)^\top,
    \end{equation}
    for $i=1\dots N$. As an immediate consequence of this structure for the gradient flow, we can state the following key result:
    
    \begin{definition}
        For any state $W\in\SSp$ of the overparameterized gradient flow \eqref{eq:gradflowOVP-def}, the \emph{invariant} is defined as $\mathscr{C}:=(\mathcal{C}_1,\dots,\mathcal{C}_{N-1})\in\left(\Reals[k\times k]\right)^{N-1}$ where
        \begin{equation}
            \label{eq:imb-def}
            \mathcal{C}_i:= W_iW_i^\top-W_{i+1}^\top W_{i+1}
        \end{equation}
    \end{definition}
    \begin{proposition}
        \label{prop:inv}
        The value of the invariant $\mathcal{C}$ along any solution of the overparameterized gradient flow \eqref{eq:gradflowOVP-def} is invariant, i.e.
        \begin{equation}
            \frac{\mbox{d}}{\mbox{d}t}\mathscr{C}=0.
        \end{equation}
    \end{proposition}

\begin{figure}
        \centering
        \includegraphics[width=0.7\linewidth]{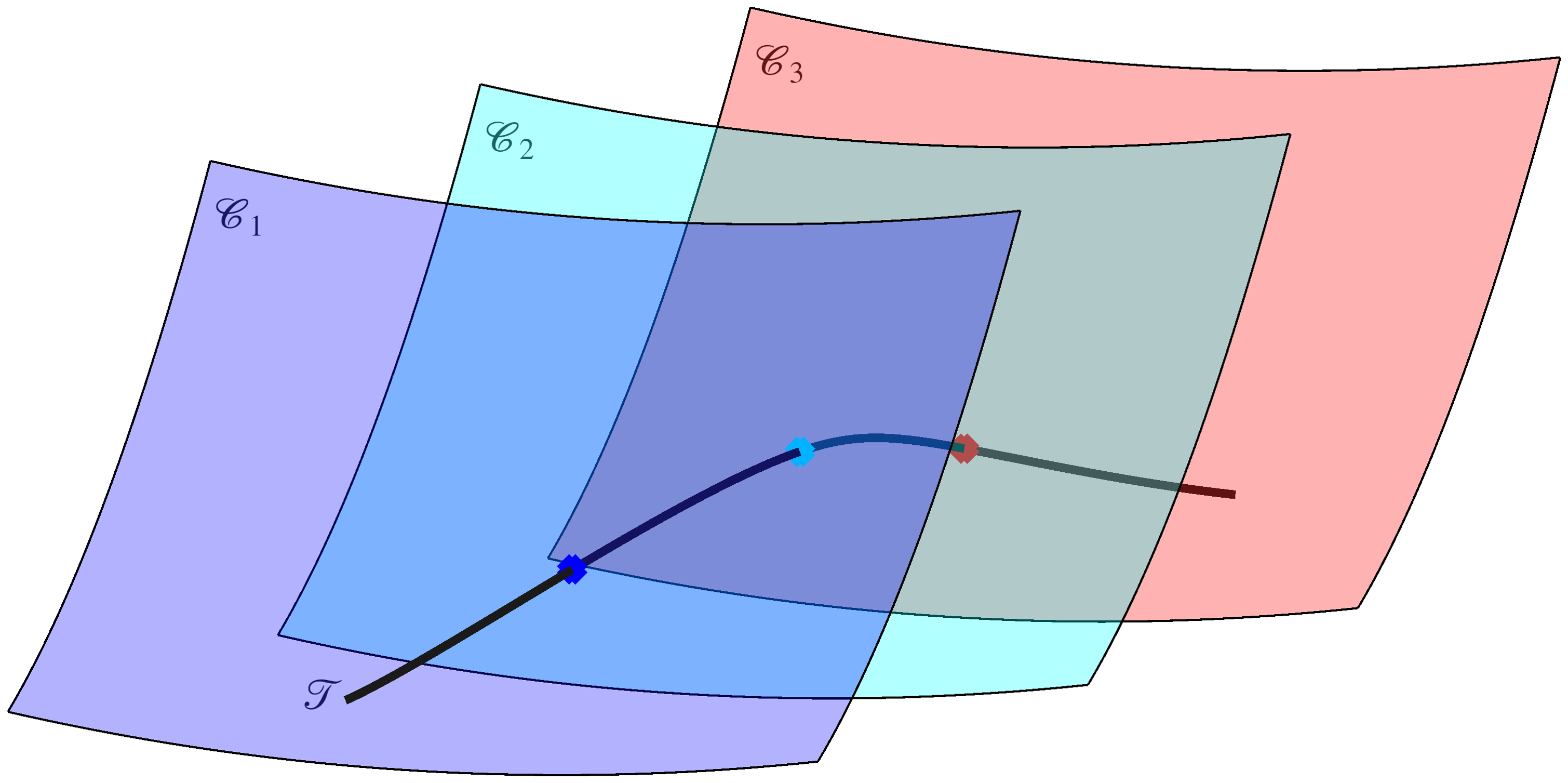}
        \caption{Illustration of the foliation of the state space of overparameterized optimization problems. The black curve illustrates a branch of the set of critical points given by $\bfW(W_1,\dots, W_N) = W^*$, and the sets $\mathscr{C}_i$ are the invariant manifold of the dynamics. Notice that the manifolds are invariant and do not intersect with each other, but every point in the parameter space is within one of such manifolds, resulting in a ``foliated'' optimization landscape. Global properties of the training can, then, be shown by simply showing that local properties in the manifold hold for all manifolds.}
        \label{fig:foliations}
    \end{figure}

    This result is independent of the cost function $f$, and is a consequence of the compositional structure assumed for overparameterization. As such, it has been noted in the literature \cite{menon2023geometry} and it has been a key result for proving the convergence of overparameterized gradient methods for different applications \cite{kawaguchi_deep_2016,chitour2018geometric,min2023convergence,de2025convergence,arora2019implicit}. This is a powerful result because it shows that the optimization landscape is ``foliated'' by invariant disjoint manifolds, as illustrated in Fig. \ref{fig:foliations}. Through this result, we obtain global convergence properties of solutions by characterizing the same property for solutions in each of these manifolds, greatly simplifying the analysis.

    Once we have established the invariant for any solution, we can leverage it to prove the following result:
    \begin{theorem}
        \label{thm:conv-gen}
        Let $f:\SSp\to\mathbb{R}$ be any real analytic and proper cost function that attains its minimum for some point in the interior of $\SSp$. Consider its overparameterized optimization problem as given in \eqref{eq:ovp-problem}. For any $W=(W_1,\dots,W_N)\in\SSp$, a solution of \eqref{eq:gradflowOVP-def} initialized at $W$ exists for all time, remains in $\SSp$, and converges to a critical point of the cost $g(W_1,\dots,W_N)$.
    \end{theorem}

    We provide the proof in the appendix for completeness. However, it has been mentioned in the literature \cite{menon2023geometry,bah_learning_2022} that pre-compactness of $\bfW(W(t))$ is enough to guarantee existence and convergence of solutions -- in this setup, properness of $f$ is enough to guarantee this condition, despite $g$ not being proper after reparameterization.

    Despite this, notice that critical points of $g$ are not necessarily critical points of $f$, as $g$ introduces multiple saddle points via orthogonality conditions of the matrix products in \eqref{eq:gradflowOVP-def}. A better condition can be stated for the case where $N=2$ as follows:

    \begin{theorem}
        \label{thm:conv-1hl}
        Let $f:\SSp\to\mathbb{R}$ be any real analytic proper cost function that attains its minimum for some point in the interior of $\SSp$. Consider its overparameterized optimization problem, as given in \eqref{eq:ovp-problem}, with $N=2$ (single hidden-layer case). Then, the overparameterized gradient-flow of \eqref{eq:gradflowOVP-def} will converge to a point in $\CSetf:=\{(W_1,W_2)\in\SSp~|~\nabla f(\bfW(W_1,W_2))=0\}$ for all but a measure-zero set of initializations. 
    \end{theorem}

    The proof of this result is given in the appendix for completeness, but it follows from a previously published result of ours \cite{de2025convergence} once it is established that for the single hidden-layer case all critical points introduced by overparameterization are \emph{strict saddles} (i.e. a point at which the gradient of the cost is zero, but the Hessian has one strictly negative eigenvalue). Being able to extend this result to a general class of functions indicates that these properties are a consequence of the overparameterized structure choice and, to some degree, independent of the specific problem being solved.



    The results above establish that overparameterization introduces a geometric structure that guarantees the convergence of gradient-flow solutions for any proper real-analytic cost function, despite the overparameterized cost no longer being proper. However, these results remain somewhat abstract, as they describe convergence in terms of invariant manifolds and precompactness rather than explicit trajectories. To gain deeper intuition, it is instructive to examine a simpler setting in which all quantities can be written in closed form.

    In the following section, we restrict attention to scalar cost functions overparameterized through linear neural networks with a single hidden layer. We call this the “vector” case, since it forces $W_1$ and $W_2$ to be vectors instead of matrices. This reduction preserves the essential compositional structure while allowing a full analytical characterization of the dynamics. Within this framework, we will explicitly describe (i) how convergence arises from the invariant geometry, (ii) how initialization imbalance affects the rate of convergence, and (iii) why these behaviors are universal across all proper analytic cost functions.

    \section{Analysis of the vector case}
    \label{sec:vectorcase}

    We consider a simplified version of the problem, where $\SSp\subseteq\Reals$ and $N=2$ -- this setup is equivalent to a single arbitrarily wide hidden layer linear neural network being trained to minimize a scalar cost. The cost is still assumed to be proper and real analytic, but no further assumptions are necessary at this point. The simpler setup allows a more complete understanding of the gradient flow behavior, as we hope to explain next.

    The overparameterized parameters are written as $w=(w_1,w_2)\in\Reals[k\times 1]\times\Reals[1\times k]$ and the parameter dynamics under gradient flow are given by
    \begin{equation}
        \begin{bmatrix}
            \dot w_1 \\ \dot w_2^\top
        \end{bmatrix} = -f'(w_2w_1)\begin{bmatrix}
            0 && 1 \\ 1 && 0
        \end{bmatrix}\begin{bmatrix}
            w_1 \\ w_2^\top
        \end{bmatrix}.
    \end{equation}
    For this setup, one can identify that the dynamics are a simple nonlinear reparameterization of a linear saddle. This fact, together with the fact that $f$ is still assumed to be proper and bounded below, allows for the following result

    \subsection{Complete characterization of convergence}
    \begin{proposition}
        \label{prop:vector-convguarantee}
        Let $\CSetf:=\{(w_1,w_2)\in\SSp~|~f'(w_2w_1)=0\}$, and let $d(w_1,w_2):=\|w_1-sw_2^\top\|$, with $s=\sign(f'(0))$. Then, any solution of the gradient flow initialized at a point $w^0=(w_1^0,w_2^0)$ that satisfies $d(w_1^0,w_2^0)>0$ will converge to a point in $\CSetf$. In particular,  $\lim_{t\to\infty}f'(w_2(t,w^0)w_1(t,w^0))=0$.
    \end{proposition}

    The proof is provided in the appendix, but it leverages properness of $f$ to show that solutions are pre-compact, and then argues that if $d(w_1^0,w_2^0)>0$ then the solutions are initialized outside the center-stable manifold of the saddle at the origin, and thus can only converge to a point where the nonlinear reparameterization $f'(\cdot)$ is zero.

    This is an important result because it indicates that the overparametrized problem can be seen as almost equivalent to the non-overparametrized problem in terms of convergence of gradient methods, except for a set of measure zero of initializations. Furthermore, the shape of this set is \emph{independent of $f$} except for the sign of its derivative at the saddle-point, indicating we can predict the bad initializations independently from the problem being solved. This indicates a high transferability of skills between problems when training a neural network, providing another important clue to justify the widespread success of neural networks.
    
    Despite that, the current results do not guarantee optimality of the solution, neither for the overparameterized formulation, nor for the original problem. Still, we will show next that ensuring optimality of the gradient-flow of $f$ is enough to do the same for $g$. For that, assume there exists a positive definite function $\alpha:\Reals\to\Reals_+$ for which $f$ satisfies
    \begin{equation*}
        |f'(w)|\geq {\alpha(f(w)-\fmin)}, \quad\quad \forall w\in\SSp.
    \end{equation*}
    A function that satisfies such a condition is said to satisfy a positive-definite (PD) Polyak-\L{}ojasiewicz inequality (\pdPLI), which is a weaker version of the Polyak-\L{}ojasiewicz inequality (\PLI) \cite{de2025remarks}. For the standard gradient flow, this condition guarantees asymptotic convergence of the cost to a global minimizer of $f$. For the overparameterized gradient flow, we can state the following corollary of Proposition \ref{prop:vector-convguarantee}:

    \begin{corollary}
        \label{cor:optguarantee}
        If $f$ satisfies a {\pdPLI} and\mbox{$f(0)>\fmin$}, then $\lim_{t\to\infty}f(w_2(t,w^0)w_1(t,w^0))=\fmin$ if and only if $d(w_1^0,w_2^0)>0$.
    \end{corollary}

    As mentioned, this result is an immediate consequence of Proposition \ref{prop:vector-convguarantee} and the fact that satisfying a \pdPLI implies that all critical points of $f$ are global minimizers. 

    \subsection{Accelerated Convergence}

    Another useful property of overparameterization is the imbalance-based acceleration of solutions. This can be characterized in general for the vector case as follows
    \begin{proposition}
        \label{prop:accelconv}
        Assume $f:\SSp\to\Reals$ is a proper, real analytic, bounded below, scalar function with minimum given by $\fmin=\min_{w\in\SSp}f(w)$. Assume further that $f(0)>\fmin$. Let $\overline{w},\widetilde{w}\in\mathbb{R}^{k\times 1}\times\mathbb{R}^{1\times k}$ be two points $\overline w = (\overline w_1,\overline w_2)$ and $\widetilde w = (\widetilde w_1,\widetilde w_2)$ such that
        \begin{itemize}
            \item $\mathbf{w}(\overline{w}):=\overline w_2\overline w_1 = \widetilde w_2\widetilde w_1=:\mathbf{w}(\widetilde{w})\in\SSp$;
            \item $c(\overline{w}):=2\mbox{trace}(\mathscr{C}(\overline{w})^2)-\mbox{trace}(\mathscr{C}(\overline{w}))^2>2\mbox{trace}(\mathscr{C}(\widetilde{w})^2)-\mbox{trace}(\mathscr{C}(\widetilde{w}))^2=:c(\widetilde{w})$;
        \end{itemize}
        then, for all $t>0$ it holds that $g(w_1(t,\overline w),w_2(t,\overline w))\leq g(w_1(t,\widetilde w),w_2(t,\widetilde w))$. If additionally we impose that $f'(\bfw(\overline w))=f'(\bfw(\widetilde w))\neq 0$, then we can strengthen the result to $g(w_1(t,\overline w),w_2(t,\overline w))< g(w_1(t,\widetilde w),w_2(t,\widetilde w))$ for all $t>0$.
    \end{proposition}

    This result is an extension of a similar result given in \cite{de2025convergence} and shows that two solutions initialized at the same ``point'' as measured by $\bfw(\cdot)$, will converge at different rates, with one being strictly faster than the other if it has a larger value of ``imbalance'' as measured by $c(\cdot)$.

    This is a known result in different overparameterized scenarios, with \cite{min2023convergence} characterizing it for any overparameterized problem whose original cost is gradient dominant. The result we present, however, is true for any proper real-analytic scalar cost, illustrating how the imbalance-based acceleration is a property of the overparameterized structure rather than of the optimization problem itself.

\section{Possible extension to sigmoidal neural networks -- A proof-of-concept}
    \label{sec:sigmoidalextension}

    \begin{figure}[t]
        \centering
        \includegraphics[width=0.45\linewidth]{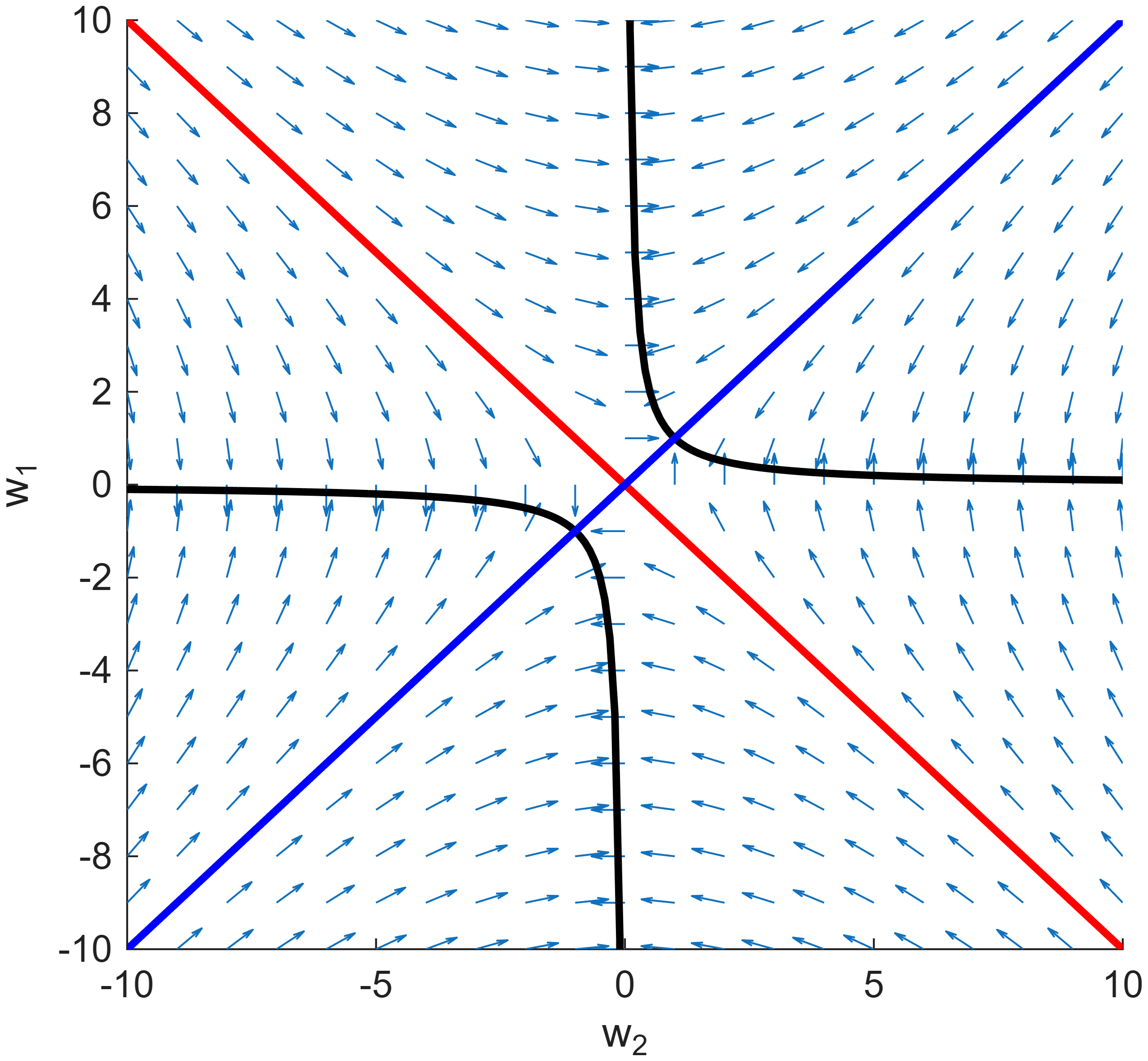}
        \includegraphics[width=0.45\linewidth]{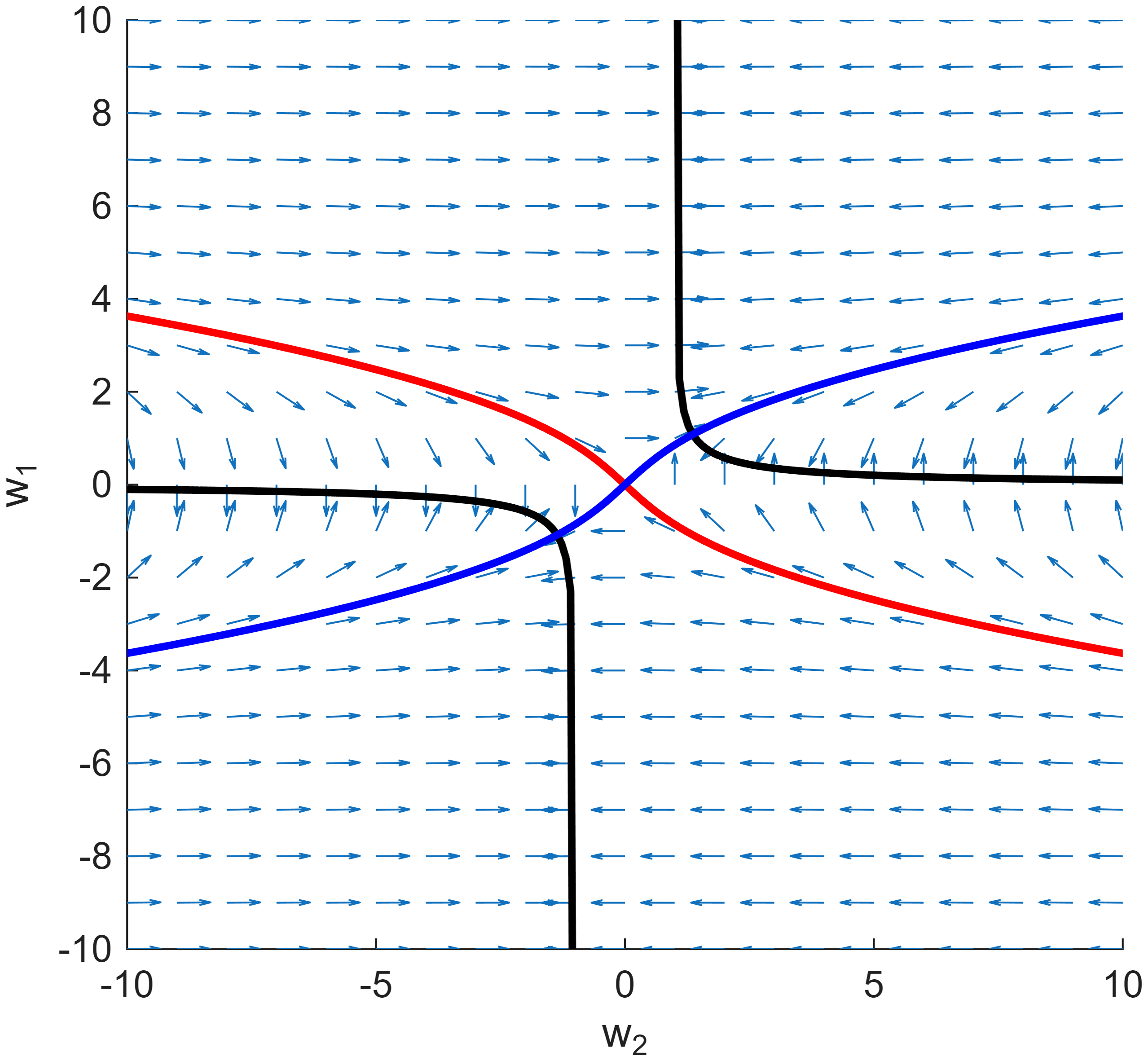}
        \caption{Illustration of the effect of sigmoidal activations to the optimization landscape of neural network training for factorization problems. The left figure displays the optimization landscape for the scalar factorization problem trained with linear neural networks, while the right one considers the same problem, but with sigmoidal networks. In solid black are displayed the target sets (global optima) for each problem; in red the center-stable manifolds of the saddle at the origin; and in blue its unstable manifold. Notice that the qualitative description of the parameter space remains unchanged: measure zero center-stable manifold for the saddle and almost everywhere convergence to the target. Despite that, notice that the center stable and unstable manifolds on the left figure segment the parameter space into four ``equivalent'' invariant subspaces, while on the right figure they segment the parameter space into four regions but of two different ``types''.}
        \label{fig:PP-linsig}
    \end{figure}

    We finish our discussion with a brief overview of how these results can be extended to sigmoidal neural networks. As a motivating example we study a scalar factorization problem. Let $f(w) = (1-w)^2$, and let the overparameterized parameters be $w_1,w_2\in\Reals$ (i.e. $k=1$). Consider the following sigmoidal activation function
    \begin{equation}
        \sigma(z) = \frac{z}{\sqrt{1+z^2}}.
    \end{equation}
    The overparameterized cost function is, then, written as $g(w_1,w_2)=(1-w_2\sigma(w_1))^2$, and the overparameterized gradient-flow becomes
    \begin{subequations}\label{eq:siggradflow}
        \begin{align}
            &\dot{w}_1:=-\frac{\partial g}{\partial w_1} = (1-w_2\sigma(w_1))\sigma'(w_1)w_2 = \frac{w_2\sqrt{1+w_1^2}-w_2^2w_1}{(1+w_1^2)^2} \\
            &\dot {w}_2 := -\frac{\partial g}{\partial w_2} = (1-w_2\sigma(w_1))\sigma(w_1) = \frac{w_1\sqrt{1+w_1^2}-w_2w_1^2}{1+w_1^2}.
        \end{align}
    \end{subequations}

    For this system, we can characterize an invariant quantity as follows:

    \begin{proposition}
        \label{prop:siginvariance}
        For the system given in \eqref{eq:siggradflow}, the following quantity is invariant along any solutions:
        \begin{equation}
            \label{eq:invsig}
            \mathscr{C}(w_1,w_2):=w_2^2-\frac{1}{2}(1+w_1^2)^2
        \end{equation}
    \end{proposition}

    The proof of this proposition is presented in the appendix, but it is purely algebraic. The more interesting consequence of this fact is that by characterizing this invariant quantity, many of the qualitative conclusions presented for linear neural networks will still hold. For example, since $\mathcal{C}$ is invariant, then one can compute $\mathcal{C}(0,0)=-1$ and find all $(w_1,w_2)$ that satisfy $\mathcal{C}(w_1,w_2)=-1$. This results in the following two curves
    \begin{equation}
        w_2=\pm w_1\sqrt{\frac{2+w_1^2}{2}}
    \end{equation}
    which characterize the center-stable and unstable manifolds of the saddle at the origin. The parallel becomes even more evident when comparing the phase portrait of this problem with and without the sigmoidal activation, both of which we present in Fig. \ref{fig:PP-linsig}.

    Notice from Fig. \ref{fig:PP-linsig} that we can also characterize necessary and sufficient conditions for optimality of the sigmoidal gradient flow solution, but also that the two phase portraits are significantly distinct, with the sigmoidal phase portrait having two distinct ``types'' of regions defined by the center-stable and unstable manifolds of the saddle, while the linear phase portrait has four equivalent regions.

\section{Conclusion}
    \label{sec:conclusion}

In this paper, we analyzed how the compositional structure of neural networks influences their optimization landscape and convergence behavior in a broad class of problems. To isolate the effects of composition from those of nonlinear activations, we focused on linear neural networks as a canonical model. We showed that the convergence guarantees traditionally associated with overparameterized linear regression/factorization extend, perhaps surprisingly, to any optimization problem whose cost function is proper and real analytic. This finding reveals that the favorable convergence behavior of deep linear models is not tied to specific data structures but is instead a structural consequence of composition itself.

We then specialized our analysis to scalar optimization problems under overparameterization, where the dynamics can be fully characterized. In this setting, we demonstrated that the center–stable manifold associated with the saddle introduced by overparameterization is universal across all proper real-analytic cost functions. Its geometry depends solely on the overparameterized structure of the parameter space and not on the particular problem instance. This universality highlights a remarkable degree of generality in neural network training: the qualitative features of the learning dynamics arise from the compositional architecture, rather than from problem-specific properties. Furthermore, we showed that the convergence rate of the overparameterized gradient flow is strongly affected by the initialization. Specifically, solutions can be significantly accelerated or slowed down depending on the level of imbalance of the initialization, as quantified by a metric introduced in this work.

Finally, we explored potential extensions to nonlinear (sigmoidal) neural networks. In the simplest nontrivial case, we showed that the qualitative structure of the parameter space persists: a strict saddle at the origin and a center–stable manifold of measure zero. We also derived an invariant quantity that remains conserved along trajectories, suggesting that the geometric and dynamical properties uncovered in the linear case may extend, to some degree, to nonlinear settings. These results provide a foundation for future work aimed at understanding how compositional structures shape not only convergence and stability but also emergent properties such as feature learning.

\acks{This work was supported by ONR Grant N00014-21-1-2431 and AFOSR Grant FA9550-21-1-0289.

The authors would also like to acknowledge Prof. Shahriar Talebi's contributions to discussions regarding sigmoidal neural networks, and our ongoing efforts to extend the results of this paper to nonlinear activations in the near future.}


\bibliography{main}

@misc{de2025remarks,
      title={Remarks on the Polyak-Lojasiewicz inequality and the convergence of gradient systems}, 
      author={Arthur Castello B. de Oliveira and Leilei Cui and Eduardo D. Sontag},
      year={2025},
      eprint={2503.23641},
      archivePrefix={arXiv},
      primaryClass={math.OC},
      url={https://arxiv.org/abs/2503.23641}, 
}

@article{de2025convergence,
  title={Convergence analysis of gradient flow for overparameterized lqr formulations},
  author={de Oliveira, Arthur Castello B and Siami, Milad and Sontag, Eduardo D},
  journal={Automatica},
  volume={182},
  pages={112504},
  year={2025},
  publisher={Elsevier}
}

@inproceedings{de2023dynamics,
  title={Dynamics and perturbations of overparameterized linear neural networks},
  author={de Oliveira, Arthur Castello B and Siami, Milad and Sontag, Eduardo D},
  booktitle={2023 62nd IEEE Conference on Decision and Control (CDC)},
  pages={7356--7361},
  year={2023},
  organization={IEEE}
}

@inproceedings{de2024remarks,
  title={Remarks on the gradient training of linear neural network based feedback for the LQR problem},
  author={De Oliveira, Arthur Castello B and Siami, Milad and Sontag, Eduardo D},
  booktitle={2024 IEEE 63rd Conference on Decision and Control (CDC)},
  pages={7846--7852},
  year={2024},
  organization={IEEE}
}

@article{chitour2018geometric,
  title={A geometric approach of gradient descent algorithms in linear neural networks},
  author={Chitour, Yacine and Liao, Zhenyu and Couillet, Romain},
  journal={Mathematical Control and Related Fields DOI:10.1007/s10107-023-01937-5},
  year={2022}
}

@article{arora2019implicit,
  title={Implicit regularization in deep matrix factorization},
  author={Arora, Sanjeev and Cohen, Nadav and Hu, Wei and Luo, Yuping},
  journal={Advances in neural information processing systems},
  volume={32},
  year={2019}
}

@inproceedings{min2021explicit,
  title={On the explicit role of initialization on the convergence and implicit bias of overparametrized linear networks},
  author={Min, Hancheng and Tarmoun, Salma and Vidal, Ren{\'e} and Mallada, Enrique},
  booktitle={International Conference on Machine Learning},
  pages={7760--7768},
  year={2021},
  organization={PMLR}
}

@inproceedings{min2023convergence,
  title={On the convergence of gradient flow on multi-layer linear models},
  author={Min, Hancheng and Vidal, Ren{\'e} and Mallada, Enrique},
  booktitle={International Conference on Machine Learning},
  pages={24850--24887},
  year={2023},
  organization={PMLR}
}

@article{bah_learning_2022,
	title = {Learning deep linear neural networks: {Riemannian} gradient flows and convergence to global minimizers},
	volume = {11},
	issn = {2049-8772},
	shorttitle = {Learning deep linear neural networks},
	url = {https://academic.oup.com/imaiai/article/11/1/307/6127129},
	doi = {10.1093/imaiai/iaaa039},
	language = {en},
	number = {1},
	urldate = {2024-02-20},
	journal = {Information and Inference: A Journal of the IMA},
	author = {Bah, Bubacarr and Rauhut, Holger and Terstiege, Ulrich and Westdickenberg, Michael},
	month = mar,
	year = {2022},
	pages = {307--353},
}

@inproceedings{eftekhari_training_2020,
	title = {Training {linear} {neural} {networks}: {Non}-{local} {convergence} and {complexity} {results}},
	shorttitle = {Training {Linear} {Neural} {Networks}},
	url = {https://proceedings.mlr.press/v119/eftekhari20a.html},
	language = {en},
	urldate = {2024-02-20},
	booktitle = {Proceedings of the 37th {International} {Conference} on {Machine} {Learning}},
	publisher = {PMLR},
	author = {Eftekhari, Armin},
	month = nov,
	year = {2020},
	note = {ISSN: 2640-3498},
	pages = {2836--2847},
}

@inproceedings{kawaguchi_deep_2016,
	title = {Deep {learning} without {poor} {local} {minima}},
	volume = {29},
	url = {https://proceedings.neurips.cc/paper/2016/hash/ f2fc990265c712c49d51a18a32b39f0c-Abstract.html},
	urldate = {2024-02-20},
	booktitle = {Advances in {Neural} {Information} {Processing} {Systems}},
	publisher = {Curran Associates, Inc.},
	author = {Kawaguchi, Kenji},
	year = {2016},
}

@inproceedings{menon2023geometry,
  title={The geometry of the deep linear network},
  author={Menon, Govind},
  booktitle={Symposium on Probability and Stochastic Processes},
  pages={1--47},
  year={2023},
  organization={Springer}
}

@article{belkin2019reconciling,
  title={Reconciling modern machine-learning practice and the classical bias--variance trade-off},
  author={Belkin, Mikhail and Hsu, Daniel and Ma, Siyuan and Mandal, Soumik},
  journal={Proceedings of the National Academy of Sciences},
  volume={116},
  number={32},
  pages={15849--15854},
  year={2019},
  publisher={National Academy of Sciences}
}

@article{bartlett2020benign,
  title={Benign overfitting in linear regression},
  author={Bartlett, Peter L and Long, Philip M and Lugosi, G{\'a}bor and Tsigler, Alexander},
  journal={Proceedings of the National Academy of Sciences},
  volume={117},
  number={48},
  pages={30063--30070},
  year={2020},
  publisher={National Academy of Sciences}
}

@article{wafi2025almost,
  title={On the (almost) Global Exponential Convergence of the Overparameterized Policy Optimization for the LQR Problem},
  author={Wafi, Moh Kamalul and de Oliveira, Arthur Castello B and Sontag, Eduardo D},
  journal={arXiv preprint arXiv:2510.02140},
  year={2025}
}

@inproceedings{du2019gradient,
  title={Gradient descent finds global minima of deep neural networks},
  author={Du, Simon and Lee, Jason and Li, Haochuan and Wang, Liwei and Zhai, Xiyu},
  booktitle={International conference on machine learning},
  pages={1675--1685},
  year={2019},
  organization={PMLR}
}

\appendix
\section{Proofs}
\subsection{Proof of Proposition \ref{prop:inv}}

This is a well-known result in overparameterized problems, having been proven in multiple other previous papers \cite{bah_learning_2022,menon2023geometry,arora2019implicit}. We include the proof here for completeness. Notice that $\dot{\mathscr{C}} = (\dot{\mathcal{C}}_1,\dots \dot{\mathcal{C}}_{N-1})$. Then, compute the time derivative of each invariant as
\begin{align*}
    \frac{\dd}{\dd t}\mathcal{C}_i = \dot W_iW_i^\top + W_i\dot W_i^\top - \dot W_{i+1}^\top W_{i+1}-W_{i+1}^\top \dot W_{i+1}
\end{align*}
and notice that 
\begin{align}
    \dot W_iW_i^\top &= -(W_N\dots W_{i+1})^\top\nabla f(\bfW)(W_{i-1}\dots W_1)^\top W_i^\top \\ &= -W_{i+1}^\top (W_N\dots W_{i+2})^\top \nabla f(\bfW)(W_i\dots W_1)^\top \\ &= W_{i+1}^\top \dot W_{i+1} 
\end{align}
and similarly that $W_i\dot W_i^\top = \dot W_{i+1}^\top W_{i+1}$. This implies that $\frac{\dd}{\dd t}\mathscr{C} = (0,\dots,0)$, completing the proof. 

\hfill$\square$

\subsection{Proof of Theorem \ref{thm:conv-gen}}

For clarity of presentation, we break this proof into a few smaller Lemmas to be proven in order. We first show that along any solution of the overparameterized gradient flow, the product of the parameter matrices is precompact. Then we show that the norm of all parameter matrices along a solution can be upper-bounded by an affine expression of the maximum singular value of the $N$-th parameter matrix. Then, using these two results we prove that the trajectory of each parameter matrix is precompact which finally allows the use of \L{}ojasiewicz's Theorem to guarantee convergence of solutions to critical points.

\begin{lemma}
    \label{lem:Wbarcomp}
    For any $W^0:=(W_1^0,\dots,W_N^0)\in\SSp$, let $W(t,W^0):=(W_1(t,W^0),\dots,W_N(t,W^0))$ be a solution to the gradient flow in \eqref{eq:gradflowOVP-def} initialized at $W^0$. Then, $\bfW(W(t,W^0))$ is pre-compact. 
\end{lemma}
\begin{proof}
    First, notice that along any solution of the gradient flow initialized in $\SSp$, the value of the overparameterized cost is non-increasing, that is
    \begin{equation}
        \frac{\dd}{\dd t}g(W(t,W^0))=-\sum_i\left\langle\nabla_{W_i}g,\dot W_i\right\rangle=-\sum_i\|\nabla_{W_i}g\|_F^2\leq 0,
    \end{equation}
    where $\langle\cdot,\cdot\rangle$ is the Frobenius inner-product. Therefore, along any trajectory, $g(W(t,W^0))\leq g(W^0)=:c$, which together with the fact that $g$ is bounded below imply that $g(W(t,W^0))$ is bounded, which tautologically implies that $f(\bfW(W(t,W^0)))$ is bounded. From this we conclude that $\bfW(W(t,W^0))$ lies in $\gW_c:=\{\bfW\in\SSp~|~f(\bfW)\leq c\}$, which is a compact set because $f$ is assumed to be proper, implying that $\bfW(W(t,W^0))$ is pre-compact.
\end{proof}

\begin{lemma}
    \label{lem:signWi}
    For any $i,j$ between $1$ and $N$, there exists a constant $c_{ij}$ such that
    \begin{equation}
        \label{eq:lema-equalitynorm}
        \|W_i\|_F^2=\|W_j\|_F^2+c_{ij}.
    \end{equation}
    Furthermore, let $\sigma_N$ be the maximum singular value of $W_N$, then it follows that one can always find $a_i$ and $b_i$ such that
    \begin{equation}
        \label{eq:lema-equalitysigma}
        \|W_i\|_F\leq a_i\sigma_N+b_i
    \end{equation}
\end{lemma}
\begin{proof}
    To begin the proof, assume $i<j$ and take the trace of both sides of \eqref{eq:imb-def} to obtain
    \begin{equation*}
        \|W_i\|_F^2=\trace{\mathcal{C}_i}+\|W_{i+1}\|_F^2,
    \end{equation*}
    which when performed iteratively results in
    \begin{equation*}
        \|W_i\|_F^2=\|W_{j}\|_F^2+\sum_{k=i}^{j-1}\trace{\mathcal{C}_k}.
    \end{equation*}
    
    The proof for $i>j$ is very similar except with a minus sign on the trace of the imbalance and appropriate summation limits. The statement is trivially true for $i=j$ and $c_{ii}=0$.

    At this point, \eqref{eq:lema-equalitynorm} is proven. To prove \eqref{eq:lema-equalitysigma} we take $i=N$ and use equivalence between the Frobenius and matrix 2 norm to justify the existence of an $a_N$ such that $\|W_N\|_F\leq a_N\sigma_N$. 
    
    With this, first notice that if $\|W_i\|_F\leq\|W_N\|_F$, then $\eqref{eq:lema-equalitysigma}$ follows with $a_i=a_N$ and $b_i=0$. 
    
    If, however, $\|W_i\|_F>\|W_N\|$, then we use $\eqref{eq:lema-equalitynorm}$ with $j=N$ to obtain that 
    \begin{equation*}
        \|W_i\|_F=\sqrt{\|W_N\|_F^2+c_{iN}}.
    \end{equation*}
    Because $\|W_i\|_F>\|W_N\|_F$, we can conclude that $c_{iN}>0$ which allows us to use the subadditive property of the square root to obtain that
    \begin{equation*}
        \|W_i\|_F\leq\|W_N\|_F+\sqrt{c_{iN}}\leq a_N\sigma_N+\sqrt{c_{iN}},
    \end{equation*}
    concluding the proof.
\end{proof}

Before proceeding to the next step of the proof, we need to formally define matrix polynomials in the context of this paper.

\begin{definition}
    For arbitrary (possibly repeating) integers $k>0$, $i_1,\dots, i_k\leq N$ and $j_0,\dots j_k$, let $A_{j_0},\dots,A_{j_k}$ be constant matrices and $X_{i_1}\dots X_{i_k}$ be variable matrices. A matrix monomial $M$ is any term of the form
    \begin{equation*}
        M(X_1,\dots,X_N) = A_{j_0}X_{i_1}A_{j_1}\dots A_{j_{k-1}}X_{i_k}A_{j_k},
    \end{equation*}
    with $k$ being the number of variable blocks $X_{i_\ell}$, and it is called the degree of the monomial. Notice that the matrices $A_i$ and $X_j$ are assumed compatible so the monomial is well-defined. A matrix polynomial is, then, written as 
    \begin{equation*}
        \mathcal{P}(X) = \sum_{\lambda\in\Lambda} a_\lambda M_\lambda(X_1,\dots,X_N)
    \end{equation*}
    and is a sum of matrix monomials as defined above, with the degree of the polynomial being the largest monomial degree of the sum.
\end{definition}

\begin{lemma}
        \label{lem:PN}
        Along any fixed trajectory, the following equality holds
        \begin{equation}
            \label{eq:WbWbTPN}
            \bfW\bfW^\top = {\left(W_NW_N^\top\right)^N + \mathcal{P}_N\left(W_N,\dots,W_2\right)},
        \end{equation}
        where $\mathcal{P}_N$ is a polynomial of degree at most $2N-2$ on the matrix variables $W_i$s and their transposes for $i\geq 2$.
    \end{lemma}

    \begin{proof}
        This lemma is proven inductively. For the first step, notice that using \eqref{eq:imb-def} for $i=1$ one can write
        \begin{align*}
            \bfW\bfW^\top &= {W_N\dots W_1W_1^\top\dots W_N^\top} \\
            &= {W_N\dots W_2\left(\mathcal{C}_1+W_2^\top W_2\right)W_2^\top\dots W_N^\top} \\
            &= {W_N\dots W_3\left(W_2W_2^\top\right)^2W_3^\top\dots W_N^\top}\;+\; \mathcal{P}_2(W_2,\dots,W_N),
        \end{align*}
        where the coefficients of $\mathcal{P}_2$ depend only on $\mathcal{C}_1$, and is of degree $2N-2$ on the variables $W_N,\dots, W_2$. Now for the induction step, assume that for some $i$ it was shown that one can write
        \begin{equation*}
            \bfW\bfW^\top = {W_N\dots W_{i+1}\left(W_iW_i^\top\right)^i W_{i+1}^\top\dots W_N^\top} + \mathcal{P}_i(W_2,\dots,W_N),
        \end{equation*}
        with $\mathcal{P}_i$ of degree at most $2N-2$ and whose coefficients depend only on $(\mathcal{C}_1,\dots,\mathcal{C}_{i-1})$. Then, apply \eqref{eq:imb-def} to obtain
        \begin{align*}
            \bfW\bfW^\top &= {W_N\dots W_{i+1} \left(W_{i+1}^\top W_{i+1}+\mathcal{C}_i\right)^iW_{i+1}^\top\dots W_N^\top} + \mathcal{P}_i(W_2,\dots,W_N).
        \end{align*}

        Notice that $(W_{i+1}^\top W_{i+1}+\mathcal{C}_i)^i$ can be expanded to $(W_{i+1}^\top W_{i+1})^i+\mathcal{R}_i(W_{i+1})$, where $\mathcal{R}_i$ collects all terms of the expansion of degree $2i-2$ or less on $W_{i+1}$ and its transpose. From this, write
        \begin{align*}
            \bfW\bfW^\top &= {W_N\dots W_{i+1} \left(W_{i+1}^\top W_{i+1}\right)^iW_{i+1}^\top\dots W_N^\top}\\&~~~~~~~+ W_N\dots W_{i+1}\mathcal{R}_i(W_{i+1})W_{i+1}^\top\dots W_N^\top+ \mathcal{P}_i(W_2,\dots,W_N)\\
            &= {W_N\dots W_{i+2} \left(W_{i+1} W_{i+1}^\top\right)^{i+1}W_{i+2}^\top\dots W_N^\top}+\mathcal{P}_{i+1}(W_N,\dots,W_2),
        \end{align*}
        where $\mathcal{P}_{i+1}$ is of degree at most $2N-2$ because it is a sum of two polynomials of degree at most $2N-2$, and has terms depending only on $(\mathcal{C}_1,\dots,\mathcal{C}_i)$. Writing the induction step expression for $i=N-1$ completes the proof.
    \end{proof}

    \begin{lemma}
        \label{lem:pn}
        There exists a polynomial $p_N:\mathbb{R}_{\ge0}\to\mathbb{R}_{\ge0}$ of degree at most $2N-2$ such that
        \[
          \|P_N(W_N,\ldots,W_2)\|_F \le p_N(\sigma_N),
        \]
        where $\sigma_N$ denotes the largest singular value of $W_N$.
    \end{lemma}
    
    \begin{proof}
        By Lemma \ref{lem:PN}, the matrix $P_N$ can be expressed as a finite sum of matrix monomials of total degree at most $2N-2$ in the variables $W_i$ and $W_i^\top$ ($i\ge2$), that is,
        \[
          P_N = \sum_{\lambda\in\Lambda} a_\lambda\, M_\lambda(W_2,\ldots,W_N),
        \]
        where each coefficient $a_\lambda\in\mathbb{R}$ and each monomial $M_\lambda$ has the form
        \[
          M_\lambda = A_{0,\lambda} X_{i_1} A_{1,\lambda} X_{i_2} \cdots A_{k-1,\lambda} X_{i_k} A_{k,\lambda},
        \]
        for some $k\le 2N-2$, with constant matrices $A_{\ell,\lambda}$ depending only on the initialization (through the invariants $\mathcal{C}_i$), and factors
        $X_{i_j}\in\{W_2,\ldots,W_N,W_2^\top,\ldots,W_N^\top\}$.
        
        Applying the triangle inequality and submultiplicativity of the Frobenius norm yields
        \[
          \|P_N\|_F \le \sum_{\lambda\in\Lambda} |a_\lambda|\, \|M_\lambda\|_F,
          \qquad
          \|M_\lambda\|_F \le \Big(\prod_{\ell=0}^{k}\|A_{\ell,\lambda}\|_2\Big)
          \prod_{j=1}^{k}\|X_{i_j}\|_F
          =: c_\lambda \prod_{j=1}^{k}\|X_{i_j}\|_F.
        \]
        
        By Lemma~10, for each $i\in\{2,\ldots,N\}$ there exist constants $a_i,b_i\ge0$ such that
        \[
          \|W_i\|_F = \|W_i^\top\|_F \le a_i\,\sigma_N + b_i.
        \]
        Hence, for every monomial $M_\lambda$ we obtain
        \[
          \|M_\lambda\|_F \le c_\lambda
          \prod_{m=2}^{N}(a_m\sigma_N+b_m)^{d_{m,\lambda}},
        \]
        where $d_{m,\lambda}$ counts how many times $W_m$ or $W_m^\top$ appears in $M_\lambda$.
        Since each monomial has total degree $\sum_{m}d_{m,\lambda}\le 2N-2$,
        the right-hand side is a polynomial in $\sigma_N$ of degree at most $2N-2$.
        
        Define
        \[
          p_N(x) := \sum_{\lambda\in\Lambda} |a_\lambda|\, c_\lambda
          \prod_{m=2}^{N}(a_m x+b_m)^{d_{m,\lambda}}.
        \]
        This is a nonnegative polynomial of degree at most $2N-2$, depending only on the constants $a_i,b_i$ and $c_\lambda$ (which in turn depend only on the initialization through the invariants).
        Combining the inequalities above gives
        \[
          \|P_N(W_N,\ldots,W_2)\|_F \le p_N(\sigma_N),
        \]
        which completes the proof.
    \end{proof}

    \begin{remark}
        \label{rem:14}
        It is evident that for a polynomial $p_1(x)$ of even degree $2k$, there exists a constant $K$ such that $p_1(x)\leq 0.5x^{2(k+1)}+K$ for all $x$ (the higher order term in $x$ dominates for large values of $x$ and the constant compensates for smaller values). We do not prove, but use this fact to prove the following lemma.
    \end{remark}

    \begin{lemma}
        \label{lem:WiCompact}
        If $\bfW(t)$ is precompact, then all $W_i(t)$ are precompact.
    \end{lemma}

    \begin{proof}
        Notice that since the Frobenius norm of a matrix upper-bounds its spectral norm, we can write
        \begin{equation*}
            \sigma_N^{2N}\leq\|(W_NW_N^\top)^N\|_F,
        \end{equation*}
        which, by using Lemma \ref{lem:PN} becomes
        \begin{equation*}
            \sigma_N^{2N}\leq \|\bfW\bfW^\top\|_F+\|\mathcal{P}_N\|_F.
        \end{equation*}
        Then, by applying Lemma \ref{lem:pn} it becomes
        \begin{equation*}
            \sigma_N^{2N}\leq \|\bfW\|_F^2+p_N(\sigma_N).
        \end{equation*}
        Then, through Remark \ref{rem:14} we know that there exists some $K$ such that $p_N(\sigma_N)\leq 0.5\sigma_N^{2N}+K$, we can write
        \begin{equation*}
            \sigma_N^{2N}\leq \|\bfW\|_F^2+0.5\sigma_N^{2N}+K
        \end{equation*}
        which in turn implies that there exist some $\eta$ and $\gamma$ such that
        \begin{equation*}
            \sigma_N\leq\eta\|\bfW\|_F^{1/N}+\gamma.
        \end{equation*}
        however, since from Lemma \ref{lem:signWi} there exist $a_i$ and $b_i$ such that $\|W_i\|_F\leq a_i\sigma_N+b_i$, it follows that there exist $\eta_i$ and $\gamma_i$ such that 
        \begin{equation*}
            \|W_i\|_F\leq \eta_i\|\bfW\|_F^{1/N}+\gamma_i.
        \end{equation*}
        Finally, since $\bfW(t)$ is precompact, then $\|\bfW(t)\|_F$ is bounded above, which in turn implies that $\|W_i(t)\|_F$ is bounded above, proving that $W_i(t)$ is precompact.
    \end{proof}

        We finally have all results to prove Theorem \ref{thm:conv-gen}. From Lemma \ref{lem:Wbarcomp}, we know that $\bfW(t)$ is precompact. Then, from Lemma \ref{lem:WiCompact}, we know that $\bfW(t)$ being precompact implies that for all $i$ between $1$ and $N$, $W_i(t)$ is precompact. From here, we can conclude the statement of the Theorem from applying Lojasiewicz's Theorem to this gradient system, since $f$ (and thus $g$) is assumed real analytic.
        
        \hfill $\square$

\subsection{Proof of Theorem \ref{thm:conv-1hl}}

To prove Theorem \ref{thm:conv-1hl}, we first state the following definition and lemma:

\begin{definition}
    Given a function $f:\SSp\to\mathbb{R}$, a point $W$ is a 
    \emph{strict saddle} of the gradient flow of $f$ if the Hessian of $f$ at $W$ has a direction of strictly negative curvature.
\end{definition}

\begin{lemma} 
    \label{lem:strictsaddles}
    If $(W_1,W_2)$ is a critical point of $g$ but $W_2W_1$ is not a critical point of $f$, then $(W_1,W_2)$ is a strict saddle of the overparameterized gradient flow dynamics.
\end{lemma}
\begin{proof}
        To begin this proof, we first show that if $(W_1,W_2)$ is a critical point of $g$, but $W_2W_1$ is not of $f$, then both $W_1$ and $W_2$ must be rank deficient. To see this, first notice that if $W_2W_1$ is not a critical point of $f$, then $\nabla f(W_2W_1)\neq 0$ by definition. However, if $(W_1,W_2)$ is a critical point of $g$, then
        \begin{align*}
            \nabla_{W_1}g(W_1,W_2):=&W_2^\top\nabla f(W_2W_1)=0\Rightarrow \textrm{Im}(\nabla f(W_2W_1)^\top)\subseteq\textrm{ker}(W_2^\top)\\
             \nabla_{W_2}g(W_1,W_2):=&\nabla f(W_2W_1)W_1^\top=0\Rightarrow \textrm{Im}(\nabla f(W_2W_1))~~\subseteq\textrm{ker}(W_1).
        \end{align*}
        Since $\nabla f(W_2W_1)\neq 0$ then $\textrm{Im}(\nabla f(W_2W_1))\neq \{0\}$ implying that the kernel of $W_2$ must also be non-empty (and similarly for $W_1$). This proves $W_1$ and $W_2$ must be rank deficient.

        Next we state the Taylor expansion of $f$ around a point $W$ in a direction $M$, which is well defined since $f$ is assumed real-analytic.
        \begin{equation*}
            f(W+M) = f(W)+f'(W)[M]+f''(W)[M,M]+o(\|M\|^2)
        \end{equation*}
        where
        \begin{align*}
            f'(W)[M]:=& \sum_{(i,j)=1}^{(n,m)} \frac{\partial f}{\partial w_{ij}}(W)m_{ij} = \langle \nabla f(W),M\rangle \\
            f''(W)[M,M]:=& \frac{1}{2}\sum_{(i,j),(k,l)}^{(n,m),(n,m)}\frac{\partial f}{\partial w_{ij}w_{kl}}(W)m_{ij}m_{kl},
        \end{align*}
        where $\langle \cdot, \cdot\rangle$ is the Frobenius inner product. From here we will build the Taylor expansion of $g$ around a point $(W_1,W_2)$ in a direction $(M_1,M_2)$. First notice that $$(W_2+M_2)(W_1+M_1)=W_2W_1+\underbrace{W_2M_1+M_2W_1}_{A}+\underbrace{M_2M_1}_{B}$$
        Where $A$ is first order in $(M_1,M_2)$ and $B$ is second order. From here we are ready to define
        \begin{align*}
            g(W_1+&M_1,W_2+M_2):=g(W_1,W_2)+g'(W_1,W_2)[(M_1,M_2)]\\&+g''(W_1,W_2)[(M_1,M_2),(M_1,M_2)]+o(\|(M_1,M_2)\|^2),
        \end{align*}
        where
        \begin{align*}
            g'(W_1,W_2)[(M_1,M_2)]:=f'(W_2W_1)[A]&\\
            g''(W_1,W_2)[(M_1,M_2),(M_1,M_2)]:=f'(W_2W_1)[B]&\\+f''(W_2W_1)[A,A]&.
        \end{align*}
        We skip the full derivation of the expression above, however notice that the $f'(W_2W_1)[B]$ term in the Hessian expression of $g$ originates from the second order chain rule. To prove that $(W_1,W_2)$ is a strict saddle of the overparameterized gradient flow, it is enough to prove that there exist $(M_1,M_2)$ such that  $$g''(W_1,W_2)[(M_1,M_2),(M_1,M_2)]<0.$$ From here, we are ready to prove the claim. 
        
        Let $\psi\in\mathbb{R}^n$ and $\phi\in\mathbb{R}^m$ be a left and right singular vector pair of $\nabla f(W_2W_1)$ associated with a nonzero singular value $\sigma>0$. 
        Next, let $\gamma$ be a unit vector and be such that $\gamma^\top W_1=0$ (no assumption is made about the value of $W_2\gamma$). Such a $\gamma$ always exists because we established that $W_1$ and $W_2$ must be rank deficient. Pick $M_1 = -\gamma\phi^\top p=\overline M_1p$ and $M_2=\psi\gamma^\top q=\overline M_2 q$ for some positive scalars $p,q$. For this choice of $M_1$ and $M_2$ notice that  $M_2W_1=0$ and thus $A=W_2M_1=W_2\overline M_1p$ and $B=M_2M_1=\overline{M}_2\overline M_1 pq$. For this choice of $M_1$ and $M_2$ we can write
        \begin{align*}
            g''(W_1,W_2)&[(M_1,M_2),(M_1,M_2)]= f'(W_2W_1)[B]+f''(W_2W_1)[A,A]\\ 
            =&pq f'(W_2W_1)[\overline M_2\overline M_1]+p^2 f''(W_2W_1)[W_2\overline M_1,W_2\overline M_1] \\ =& apq + bp^2.
        \end{align*}
        From the chosen $M_1$ and $M_2$ notice that
        \begin{align*}
            a &= f'(W_2W_1)[\overline M_2\overline M_1] \\&= \trace{\nabla f(W_2W_1)^\top \overline M_2 \overline M_1} \\&= \trace{\nabla f(W_2W_1)^\top (-\psi\phi^\top)}\\&= -\psi^\top\nabla f(W_2W_1)\phi \\&=  -\sigma,
        \end{align*}
        however the sign of $b$ is undefined (and if $W_2\gamma=0$, then $b=0$) although we know it is finite due to continuity of $f''$. Despite that, if we pick the particular case where $p=q^2$ then we can write
        \begin{equation*}
            g''(W_2W_1)[(M_1,M_2),(M_1,M_2)]=-\sigma q^3+bq^4.
        \end{equation*}
        From the expression above, let $\overline q=\sigma/(2|b|)$ if $b\neq 0$ or any positive value if $b=0$, then for all $0<q<\overline q$, $-\sigma q^3+bq^4<0$. 
        
        Hence, there exists a direction $(M_1,M_2)$ along which the Hessian of $g$ is strictly negative, proving that every $(W_1,W_2)$ that consists on a critical point of $g$ but that is such that $W_2W_1$ is not a critical point of $f$ is a strict saddle.
    \end{proof}

We also state the following Theorem, and refer the reader to \cite{de2025convergence} for the proof.

\begin{theorem}
    \label{thm:automaticaconv}
    Consider a nonlinear system of the form $\dot x = f(x)$, $f:\mathcal{X}\to\mathcal{TX}$. Suppose that $E\subseteq \mathcal{X}$ is a set consisting of strict saddle
    equilibria of the system.
    Then the set $\mathcal{C}_E$ of points $x_0\in \mathcal{X}$ whose trajectories
    converge to points in $E$ has measure zero.
\end{theorem}

Finally, we know that all solutions converge to a critical point due to Theorem \ref{thm:conv-gen}. Then, Lemma \ref{lem:strictsaddles}  tells us that all points in $$E=\{(W_1,W_2) ~|~ \nabla_{W_{1,2}}g(W_1,W_2)=0,~\nabla f(W_2W_1)\neq 0\},$$ are strict saddles, which allows us to invoke Theorem \ref{thm:automaticaconv} to prove Theorem \ref{thm:conv-1hl}\hfill $\square$

\subsection{Proof of Proposition \ref{prop:vector-convguarantee}}

Let $z\coloneqq w_2 w_1$ and $g(w)\coloneqq f(z)$ for $w=(w_1,w_2)\in\mathbb{R}^k\times\mathbb{R}^k$.
For the gradient flow $\dot w=-\nabla g(w)$ we have
\[
\nabla g(w)=f'(z)\!\begin{bmatrix}w_2\\ w_1\end{bmatrix},\qquad
\dot w_1=-f'(z)\,w_2,\quad \dot w_2=-f'(z)\,w_1.
\]
Define
\[
S(t)=\|w_1\|^2+\|w_2\|^2,\qquad z(t)=w_2 w_1,\qquad V(t)=f(z(t)).
\]
Then
\[
\dot z=-S\,f'(z),\qquad \dot S=-4z\,f'(z),\qquad \dot V=f'(z)\dot z=-S\,[f'(z)]^2\le 0.
\]
The quantity $D\coloneqq S^2-4z^2$ is conserved: $\dot D=0$ and thus $D(t)\equiv D(0)\ge 0$.
Introduce
\[
u\coloneqq w_1-s\,w_2,\qquad v\coloneqq w_1+s\,w_2,\qquad 
\]
In these coordinates,
\[
\dot u=s\,f'(z)\,u,\qquad \dot v=-\,s\,f'(z)\,v,
\]
and
\[
S=\tfrac12\big(\|u\|^2+\|v\|^2\big),\qquad
z=\tfrac{s}{4}\big(\|v\|^2-\|u\|^2\big),\qquad
D=\|u\|^2\,\|v\|^2
\]

Let $\mathcal S^+=\{f'(z)>0\}$ and $\mathcal S^-=\{f'(z)<0\}$; their common boundary is $Z=\{f'(z)=0\}$, we show they are forward invariant. Suppose towards contradiction a trajectory starting in $\mathcal S^+$ were to enter $\mathcal S^-$ at the first time $t_\ast>0$. Then
$f'(z(t_\ast))=0$ and hence $\dot w(t_\ast)=0$; by uniqueness, the solution with initial
condition $w(t_\ast)$ is constant, thus the trajectory cannot cross the boundary. The same holds for the
roles of $\mathcal S^\pm$ reversed. Thus $\mathcal S^\pm$ are forward invariant. In particular, while
$f'(z(t))\neq 0$ the sign of $\dot z=-S f'(z)$ is fixed and $z(t)$ is strictly monotone. Similarly, the sets $\{u=0\}$ and $\{v=0\}$ are forward invariant because $\dot u=s f'(z)\,u$ and
$\dot v=-s f'(z)\,v$. Note $\{u=0\}$ coincides with $\Delta_+=\{w_1=w_2\}$ if $s=+1$ and with
$\Delta_-=\{w_1=-w_2\}$ if $s=-1$; the correspondence is reversed for $\{v=0\}$. Since $\dot V\le 0$, we have $V(t)\le V(0)$ for all $t$. By the properness of $f$, the set $\{\,\xi:\ f(\xi)\le V(0)\,\}$ is compact, so $z(t)$ remains bounded. Then $S^2(t)=4z^2(t)+D(0)$ bounds $S(t)$, and hence $(w_1(t),w_2(t))$ is bounded. Thus all trajectories are precompact.

By the real analyticity of $g(w)=f(w_2^\top w_1)$), \L{}ojasiewicz’s 
theorem says that every precompact trajectory of $\dot w=-\nabla g(w)$ converges to a single critical point of $g$. The critical points satisfy $\nabla g(w)=0$, i.e.,
\[
f'(z)=0 \quad\text{or}\quad (w_1,w_2)=(0,0).
\]
Recall $d(w_1,w_2)=\|u\|$. Consider the following cases. First, suppose $D(0)>0$. Then $\|u(0)\|>0$ and $\|v(0)\|>0$, and since $D(t)=\|u(t)\|^2\|v(t)\|^2\equiv D(0)$, both $\|u(t)\|$ and $\|v(t)\|$ stay bounded away from zero for all $t$. Hence the limit cannot be $(0,0)$, so the trajectory
converges to a point in $Z$. Then suppose $D(0)=0$ and $d(w_1^0,w_2^0)>0$. Here $\|u(0)\|>0$ forces $\|v(0)\|=0$, and by invariance $v(t)\equiv 0$ for all $t$. Along this invariant line we have $z=-\tfrac{s}{4}\|u\|^2$ and $\dot u=s\,f'(z)\,u$.
If the trajectory were to converge to $(0,0)$, then $u(t)\to 0$ and hence $z(t)\to 0$, so for $t$ large
we would have $s\,f'(z(t))>c>0$ (because $f'(0)=s$). But then
\[
\frac{d}{dt}\|u(t)\|^2 \;=\;2\,s\,f'\big(z(t)\big)\,\|u(t)\|^2 \;\ge\; 2c\,\|u(t)\|^2,
\]
which does not allow $\|u(t)\|\to 0$ forward in time. Thus the origin cannot be the limit. By \L{}ojasiewicz's theorem, the trajectory must converge to a point with $f'(z)=0$, i.e., to an element of $Z$. In all cases compatible with $d(w_1^0,w_2^0)>0$, the trajectory converges to a point in $Z$.
Consequently,
\[
\lim_{t\to\infty} f'\!\big(w_2(t)^\top w_1(t)\big)=0.
\]
\hfill$\square$

\subsection{Proof of Corollary \ref{cor:optguarantee}}

By assumption, $|f'(0)|\ge \alpha\big(f(0)-f_{\min}\big)>0$, so $f'(0)\neq 0$. Set $s=\operatorname{sign}(f'(0))$, and write $z(t)=w_2(t)^\top w_1(t)$ and $V(t)=f(z(t))$. If $d(w_1^0,w_2^0)>0$, Proposition \ref{prop:vector-convguarantee} yields $(w_1(t),w_2(t))\to (w_1^\ast,w_2^\ast)$ with $f'\!\big(w_2^{\ast\top}w_1^\ast\big)=0$. From assumption of \pdPLI, $f'(z)=0$ implies $f(z)=f_{\min}$, hence $V(t)\to f_{\min}$ by continuity. For the converse, argue by contrapositive and suppose $d(w_1^0,w_2^0)=0$. Let $u=w_1-s\,w_2$. From $\dot w_1=-f'(z)\,w_2$ and $\dot w_2=-f'(z)\,w_1$ we obtain $\dot u=s\,f'(z)\,u$, hence $u(0)=0$ implies $u(t)\equiv 0$ and therefore $w_1(t)=s\,w_2(t)$ for all $t$. As in Proposition~5, the trajectory is precompact (since $\dot V=-S[f'(z)]^2\le 0$ and $f$ is proper), and $g(w)=f(w_2^\top w_1)$ is real-analytic; therefore, by \L{}ojasiewicz's theorem, the trajectory converges to a single critical point $w^\ast$ of $g$. Critical points of $g$ satisfy either $f'(z^\ast)=0$ or $w^\ast=(0,0)$. On the invariant line $\{u=0\}$ we have $z=s\,\|w_1\|^2$ and thus $s\,z\ge 0$. We claim there is no point on the ray $\{s\,z>0\}$ with $f'(z)=0$. Indeed, define $g_1(\rho)=f(s\rho)$ for $\rho\ge 0$. Then $g_1'(0)=s f'(0)>0$. If there were $\rho_\ast>0$ with $g_1'(\rho_\ast)=0$, the inequality would force $g_1(\rho_\ast)=f_{\min}<g_1(0)$. By the mean value theorem there exists $\hat\rho\in(0,\rho_\ast)$ with $g_1'(\hat\rho)<0$, so by continuity $g_1'$ would vanish first at some $\rho_0\in(0,\rho_\ast)$ with $g_1(\rho_0)>f_{\min}$, contradicting the \pdPLI assumption (which gives $g_1'=0\Rightarrow g_1=f_{\min}$). Hence $f'(z)\neq 0$ for all $s\,z>0$, which excludes critical points of the form $f'(z)=0$ on the stable manifold of the saddle. Therefore the only critical point accessible on $\{u=0\}$ is $w^\ast=(0,0)$, so the \L{}ojasiewicz theorem yields $(w_1(t),w_2(t))\to(0,0)$; in particular, $z(t)\to 0$ and $V(t)=f(z(t))\to f(0)>f_{\min}$.  Therefore, $\lim_{t\to\infty} f\big(w_2(t)^\top w_1(t)\big)=f_{\min}$ holds if and only if $d(w_1^0,w_2^0)>0$.\hfill$\square$

\subsection{Proof of Proposition \ref{prop:accelconv}}

    Remember that $\bfw(w_1,w_2)=w_2w_1$, and drop the explicit dependencies when it is obvious by context. Notice from the gradient flow that
    \begin{align}
        \frac{\mbox{d}}{\mbox{d}t}\bfw&=-f'(\bfw)(w_2w_2^\top+w_1^\top w_1)\nonumber \\ &= -f'(\bfw)(2w_2w_2^\top+\trace{\mathscr{C}})\nonumber\\
        &=-f'(\bfw)\sqrt{c+4\mathbf{w}^2} \label{eq:prop7exp} \\
        \frac{\mbox{d}}{\mbox{d}t}f(\bfw)&=-[f'(\bfw)]^2\sqrt{c+4\mathbf{w}^2}
    \end{align}
    where $c:=2\trace{\mathscr{C}^2}-\trace{\mathscr{C}}^2$. From the expressions above we can see that the imbalance constant $c$ merely rescales the rate of $\bfw$ and $g$.

    Let $\tau(t)$ be a time reparameterization such that $\frac{\dd \tau}{\dd t}=\sqrt{c+4\bfw(\tau)^2}$ with $\tau(0)=0$. Then $\frac{\dd \bfw}{\dd \tau}=-f'(\bfw)$ and $\frac{\dd f}{\dd \tau}=-[f'(\bfw)]^2$. Therefore, $\bfw(\tau)$ and $f(\bfw(\tau))$ will have the same trajectory for the same values of $\tau$ if initialized at the same point, independently of the value of the imbalance $c$.

    Then, to prove the statement it is enough to show that for two points $\overline w$ and $\widetilde w$ that satisfy the conditions in the proposition, $\overline \tau>\widetilde \tau$ for all $t>0$, since $\frac{\dd}{\dd t}f\leq 0$.

    To prove that, let $\Delta(t):=\overline \tau(t)-\widetilde \tau(t)$. Notice that $\Delta(0)=0$ and $\dot\Delta(0)=\sqrt{\overline c+4\overline\bfw^2}-\sqrt{\widetilde c+4\widetilde\bfw^2}>0$, and furthermore notice that for any $\bar t$ such that $\Delta(\bar t)=0$ then $\dot\Delta(\bar t)>0$ necessarily. We argue that this implies that $\Delta(t)>0$ for all $t>0$. To see that, first notice that $\Delta(\epsilon)>0$ for some $\epsilon>0$ since $\dot\Delta(0)>0$. Then, assume for contradiction that at some time $\bar t>0$, $\Delta(\bar t)=0$. In particular, let $\bar t$ be the smallest $t>0$ for which this condition is satisfied. Since $\dot\Delta(\bar t)>0$, then there must exist some $h\in(\epsilon,\bar t)$ for which $\Delta(h)<0$, however if that is the case, then $\Delta(\epsilon)>0$ and $\Delta(h)<0$ which by the mean value theorem implies that there must exist some $\delta\in(\epsilon,h)$ such that $\Delta(\delta)=0$, which breaks the condition that $\bar t$ is the smallest time for which $\Delta(\bar t)=0$, reaching contradiction. 

    This proves that $\Delta(t)>0$ for all $t>0$ which in turn implies that $\overline \tau(t)>\widetilde \tau(t)$, which implies that $f(t,\overline w)<f(t,\widetilde w)$ for all $t>0$.\hfill $\square$

\subsection{Proof of Proposition \ref{prop:siginvariance}}

To prove this proposition, all one needs to do is to algebraically compute $\frac{\dd}{\dd t}\mathscr{C}$ as follows
\begin{align*}
    \frac{\dd}{\dd t}\mathscr C &= 2w_2\dot w_2-2(1+w_1^2)w_1\dot w_1 \\ 
    &= 2w_2w_1\left(\frac{\sqrt{1+w_1^2}-w_2w_1}{1+w_1^2}-(1+w_1^2)\frac{\sqrt{1+w_1^2}-w_2w_1}{(1+w_1^2)^2}\right) \\ &=0.
\end{align*}
\hfill $\square$

\end{document}